# Continual Learning for Steganalysis

Zihao Yin[1)]    Ruohan Meng[1) 3)]    Zhili Zhou*[2)1)]

[1)] Engineering Research Center of Digital Forensics, Ministry of Education, Nanjing University of Information Science and Technology, 210044, Nanjing, China

[2)] Institute of Artificial Intelligence and Blockchain, Guangzhou University, Guangdong, 510006, China

[3)] School of Computer Science Engineering, Nanyang Technological University, 639798, Singapore

**Abstract**. To detect the existing steganographic algorithms, recent steganalysis methods usually train a Convolutional Neural Network (CNN) model on the dataset consisting of corresponding paired cover/stego-images. However, it is inefficient and impractical for those steganalysis tools to completely retrain the CNN model to make it effective against both the existing steganographic algorithms and a new emerging steganographic algorithm. Thus, existing steganalysis models usually lack dynamic extensibility for new steganographic algorithms, which limits their application in real-world scenarios. To address this issue, we propose an accurate parameter importance estimation (APIE) based-continual learning scheme for steganalysis. In this scheme, when a steganalysis model is trained on the new image dataset generated by the new steganographic algorithm, its network parameters are effectively and efficiently updated with sufficient consideration of their importance evaluated in the previous training process. This approach can guide the steganalysis model to learn the patterns of the new steganographic algorithm without significantly degrading the detectability against the previous steganographic algorithms. Experimental results demonstrate the proposed scheme has promising extensibility for new emerging steganographic algorithms.

## 1 INTRODUCTION

Steganography is the technique that imperceptibly hides secret information into a multimedia carrier [1]–[8]. As an adversary of steganography, steganalysis aims to determine the existence of secret information hidden in a multimedia carrier.

Image steganalysis methods can be roughly divided into two categories: traditional machine learning (ML)-based methods [9]–[14] and the deep learning (DL)-based methods [15]–[21]. Generally, ML-based methods manually extract image statistic features, and use a trained binary classifier to detect whether a given image is a stego-image or not. With the development of deep learning, researchers explored deep learning techniques to improve detection accuracy by jointly optimizing the image features as well as the classifier. In 2015, Qian et al. [15] proposed Gaussian-Neuron CNN to automatically learn effective features for the steganalysis task. Subsequently, Xu et al. [16] proposed XuNet, which employed the absolute value layer and TanH activation in the front part of the network. It was the first model that obtains competitive performance compared with the ML models. In 2019, Boroumand et al. [21] proposed SRNet, which is a complete end-to-end model, including no fixed preprocessing layers. During the training, the network can automatically learn optimal filters to extract steganographic features. Recently, Jia et al. [22] proposed a consensus-clustering-based automatic distribution matching scheme, called CADM, which can automatically match inconsistent distributions in cross-domain scenarios for steganalysis.

To detect the existing steganographic algorithms, a well-designed steganalysis model is usually trained on a dataset consisting of paired cover/stego-images generated by those steganographic algorithms. It is notable that, when a new steganographic algorithm emerges, the steganalysis model is also expected to be effective for the new emerging steganographic algorithm. To this end, two model training strategies, i.e., fine-tuning [23], [24] and jointly training, are popularly adopted. If a steganalysis model is fine-tuned on the dataset generated by the new steganographic algorithm, the fine-tuned steganalysis model could perform well against the new steganographic algorithm. However, the detection performance of retrained model will degrade significantly against the previous steganographic algorithms, due to the phenomenon known as catastrophic forgetting problem [25]. Also, it is inefficient and impractical for those steganalysis tools to completely retrain the steganalysis model on both the previous dataset and the new dataset.

Therefore, as new steganographic algorithms continuously emerge, the above steganalysis models show limited extensibility, which makes them hard to be applied in real-world scenarios.

Recently, continual learning has become a promising paradigm that can address the catastrophic forgetting problem when learning a neural network model for a new task. In the existing continual learning schemes [26]–[31], the Memory Aware Synapses (MAS) [30] is one of most typical regularization addition-based continual learning schemes. It updates the parameters of neural network model according to the parameter importance on the previous task, when learning for a new task. Inspired by MAS, to address the problem of catastrophic forgetting in steganalysis, we propose an accurate parameter importance estimation (APIE)-based continual learning scheme for steganalysis. In this scheme, the parameter importance of steganalysis model is estimated relying on the curvature of output function, and the parameter importance weights across multiple tasks are accumulated to obtain the regularization term. Consequently, the proposed APIE-based continual learning scheme can be extended well to new steganographic algorithms in an effective and efficient way, and thus performs well on new steganographic algorithms while maintaining satisfactory performance on previous ones.

The major contributions of this paper are summarized as follows:

1) It is the first attempt to explore the continual learning on steganalysis, which aims to extend steganalysis model to new emerging steganographic algorithms in an effective and efficient manner.

2) An APIE-based continual learning scheme is proposed. In this scheme, the parameter importance of steganalysis model is estimated relying on the curvature of output function, and the parameter importance weights across multiple tasks are accumulated to obtain the regularization term. By accurately estimating the parameter importance, the framework modifies model to mitigate catastrophic forgetting for steganalysis. Experimental results demonstrate the proposed scheme has promising extensibility for steganographic algorithms.

## 2 The Proposed Method

For an extensible steganalysis model, we design an accurate parameter importance estimation (APIE)-based continual learning framework, as shown in Fig. 1. The popular steganalysis model, *i.e.*, SRNet, is used in the proposed framework. A steganography dataset generated by each steganography algorithm is considered as a task. The key idea of this continual learning framework is to estimate the importance of network parameters based on past learned tasks. When training on a new task, the parameters of steganalysis model are updated with sufficient consideration of their importance evaluated on the previous tasks. Where, the more important the parameter is, the less it is updated. In the following, we will first introduce MAS, and then describe the proposed continual learning scheme for steganalysis.

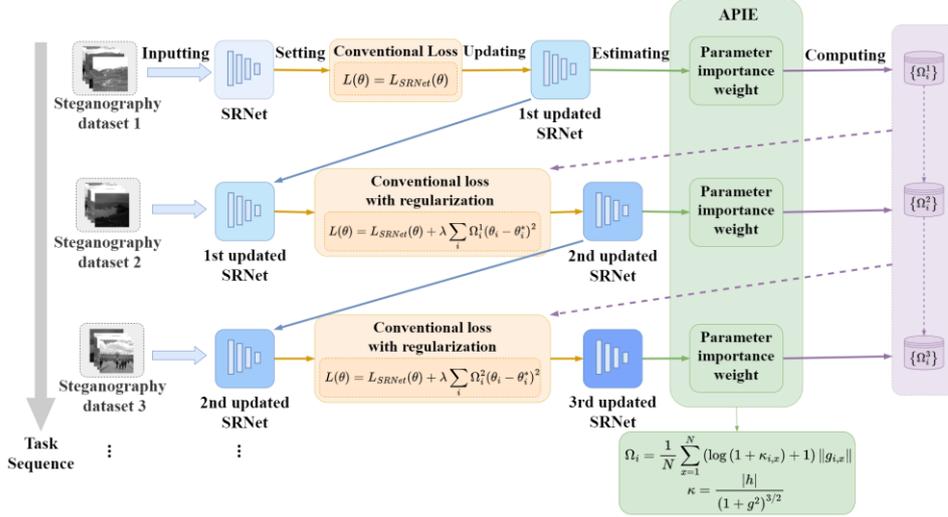

Fig. 1. Overview of our continual learning approach for steganalysis. During each steganalysis task iteration, the model estimates importance weights for each network parameter and stores them. After each task (Steganography Dataset), the model calculates regularization term by using the stored importance weights and combines it with task-specific backward loss. The model is updated under the restriction of the regularization term in order to obtain favorable performance on new task while maintaining satisfactory performance on previous tasks.

### 2.1 Preliminary

The regulation addition-based continual learning method, i.e., MAS, is first briefly described. After each training process of the task, the MAS method estimates an importance weight for each network parameter indicating the importance of this parameter to the previously learned task. It estimates the importance weight by computing the sensitivity of the learned function $F$ to the parameters changes:

$$F(x_k; \theta + \delta) - F(x_k; \theta) \approx \sum_i g_i(x_k)\delta_i \quad (1)$$

$$\Omega_i = \frac{1}{N}\sum_{k=1}^{N}\|g_i(x_k)\| \quad (2)$$

where $x_k$ is the $N$ samples from the previous task, $\delta_i$ a small change to model parameter $\theta$, $g_i(x_k) = \frac{\partial F(x_k)}{\partial \theta_i}$ the gradient of the learned function with respect to the parameter $\theta_i$ evaluated at the data point $x_k$, and $\Omega_i$ the importance weight of parameter $\theta_i$. When learning a new task, a regularization term is added to penalize any changes to important parameters:

$$L(\theta) = L_n(\theta) + \lambda \sum_i \Omega_i (\theta_i - \theta_i^*)^2 \quad (3)$$

Where, $\theta^*$ is the parameter value determined by the optimization for the previous task in the sequence, $L(\theta)$ the task-specific backward loss and $\theta_i$ is the current parameter value during training, and the hyperparameter $\lambda$ a positive real number representing the weight factor of the regularization term. After each task, the importance weight $\Omega_i$ is computed by accumulating all the previous estimated values.

### 2.2 The Proposed APIE-Based Continual Learning for Steganalysis

#### A. Motivation

The main challenge of regularization addition-based continual learning methods is how to accurately identify the important parameters and minimize the negative effects of parameter consolidation on the network learning capacity. Although the MAS method shows more promising performance than the other regularization-based methods, it still has two main shortcomings. One is that it is not accurate enough to estimate the parameter importance by only using the gradient of output function. The other is directly accumulating importance weight cross multiple tasks will dilute the impacts of parameters that are important for previous task, leading to unsatisfactory performance after training. Therefore, we propose an accurate parameter importance estimation (APIE)-based MAS method. In this method, the importance weight of network parameters is computed based on curvatures of output function, and Peak-Weight algorithm is proposed to maintain the privilege of important parameter.

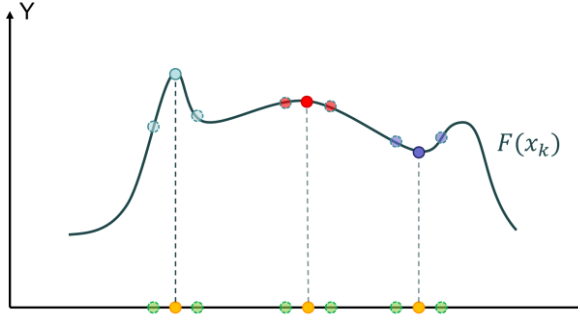

Fig. 2. The illustration of MAS estimates the parameter importance based on gradient, i.e., the sensitivity of the output function to parameter changes. The gradient of output $Y$ with respect to the parameter $\theta_i$ evaluated at a given data point $x_k$ is used to measure the importance of that parameter.

B. Curvature-based Weight Importance Computation

MAS estimates the parameter importance based on the gradient of the learned function with respect to the parameter. As shown in Fig. 2, parameters with the same gradient (in yellow) have nonequivalent influence on the function output when the parameters are slightly changed (in green). Since the change degrees of parameters are based on their importance in current task, it is a key issue to estimate their importance more accurately. Thus, it is required to use a larger field of view to evaluate the sensitivity of the output function to parameters. In other words, the smoothness of the function near the data point needs to be also considered. Therefore, we attempt to use the curvatures to describe the parameters importance by

$$\kappa_i(x_k) = \frac{|h_i|}{(1 + g_i^2)^{3/2}} \qquad (4)$$

where $h_i(x_k) = \frac{\partial^2 F(x_k)}{\partial \theta_i^2}$ is the second gradient of the output function with respect to the parameter $\theta_i$. Considering the curvature with the gradient, the importance of each parameter is estimated by

$$\Omega_i = \frac{1}{N} \sum_{k=1}^{N} C_i(x_k) \qquad (5)$$

$$C_i(x_k) = (log(1 + \kappa_i) + 1)\|g_i\| \qquad (6)$$

Where, the output function $F$ is multi-dimensional, as is the case for most neural networks. Considering the computational efficiency of calculating the second gradients for each output, we use the second gradients of the squared $l_2$ norm of the output function, i.e., $h_i(x_k) = \frac{\partial^2 [l_2^2(F(x_k))]}{\partial \theta_i^2}$.

C. Peak-Weight

As mentioned above, the original MAS method directly accumulates the currently estimated $\Omega_i$ after each task. Parameters with great importance for one task may have less impact in other tasks. Thus, after the training over a sequence of tasks, the accumulated importance weight will dilute the privilege of these parameters for specified task, leading to unsatisfactory performance for overall performance.

To tackle such limitation, we propose Peak-Weight algorithm, adopting $\Omega_{ad}$ to replace $\Omega_i$, to maintain the influence of parameters that have great impact in specified task by

$$\Omega_{ad,i} = \alpha \Omega_{peak,i} + \beta \overline{\Omega_i} \qquad (6)$$

where $\Omega_{peak,i}$ is the maximum one of $\Omega_i$ in the past sequence tasks and $\overline{\Omega_i}$ refers to the mean value of $\Omega_i$ for the past tasks. The hyperparameters $\alpha$ and $\beta$ are used to balance the impact of these two values.

3 Experiment

3.1 Dataset and Experimental Platform

The dataset adopts the BOSSBase v1.01 [6] database. This dataset is widely used for evaluating the experiments of steganography and steganalysis, and includes 10,000 grayscale PGM images of size 512×512. The image database is splatted into 60% for training, 20% for validation, and 20% for test. The cover images are first resized from 512 × 512 to 256 ×

256, and then four steganography algorithms, i.e., WOW [5], S-UNIWARD [3], HILL [7], and UTGAN [4], are used to generate corresponding four steganographic datasets. The steganography algorithms are implemented with embedding rates of 0.4 bpp (bits per pixel). All our experiments are performed on NVIDIA rtx3090 GPU platform.

3.2 Implement Details

We regard the steganalysis on each dataset an individual task, and train the model sequentially. Specifically, the model is trained on WOW (Task 1), S-UNIWARD (Task 2), HILL (Task 3) and UTGAN (Task 4) sequentially. For training the network, we set the number of epochs as 80. Due to the limitation of GPU memory, the batch size of training is set as 32. To optimize the network, the SGD algorithm is used to update the network parameters, and the initial value of learning rate is 0.01. However, for the learning rate, it is found that it is more suitable to reduce it to one-fifth when training for the second tasks. That is because we need to fine-tune the network on these additional tasks, that is, the weights should not change significantly for these incremental tasks. In addition, the forgetting regularization hyperparameter $\lambda$ in our proposed framework is independent between different layers of the network. As SRNet can be functionally divided into two segments: the first seven layers of the network responsible for extracting the noise residuals, and the later five layers used for compacting the feature maps and classification. Thus, the hyperparameter $\lambda$ of these two segments is assigned to different initial values in our experiment, *i.e.*, 1.2 and 1, respectively.

3.3 Results on Benchmark Datasets

A. Baseline Setup

The baseline is the method that is sequentially and independently feeding above steganography datasets into a model for training without continual learning. In the setting of baseline, due to the catastrophic forgetting, the weights well-trained on the previous dataset are covered and updated by the new steganography dataset, thus neglecting the previous performance. After training on the last dataset, we evaluate the ultimate model on all the test datasets.

TABLE I

DETECTION ACCURACIES (%) OF BASELINE, REFERENCE AND OUR APPROACH ON FOUR SEQUENTIALLY STEGANOGRAPHIC METHODS

| Embedding Method | Baseline | Proposed | Reference |
|---|---|---|---|
| WOW | 77.16 | 83.20 | 91.73 |
| S-UNIWARD | 74.95 | 79.45 | 89.15 |
| HILL | 80.52 | 85.80 | 88.83 |
| UTGAN | 85.48 | 81.65 | 86.43 |

B. Comparison with Baselines

Table I shows the comparison between the baselines and our approach. The model is trained sequentially on tasks using schemes of baseline and proposed method respectively. Reference method refers to the results of the model trained on each dataset individually from scratch. The other results shown in this table are the performance of the ultimate model after training all tasks on test datasets. It shows that the parameter consolidation on the network has a certain degree of negative effects on learning new tasks, making the proposed approach slightly underperform on the last task compares with baseline. However, it has significant improvements of detection accuracies on all rest of last tasks. It clearly indicates that the proposed scheme is capable of obtaining satisfactory results on new emerging steganographic algorithm while maintaining acceptable performance on the previous one. That proves the proposed scheme has good extensibility for steganographic algorithms.

3.4 Ablation Studies

In this section, we conduct an ablation study to illustrate the validity of the proposed Curvature-based estimation algorithm and Peak-Weight algorithm used in our approach. We adopt the original regulation addition-based continual learning scheme, *i.e.*, MAS, as the baseline method, and APIE-continual learning scheme to observe the performance gain, which is detailed in Fig. 3. It can be clearly observed that the proposed APIE-continual learning scheme outperforms the original MAS significantly. The ablation analysis results show the two proposed algorithm, i.e.,

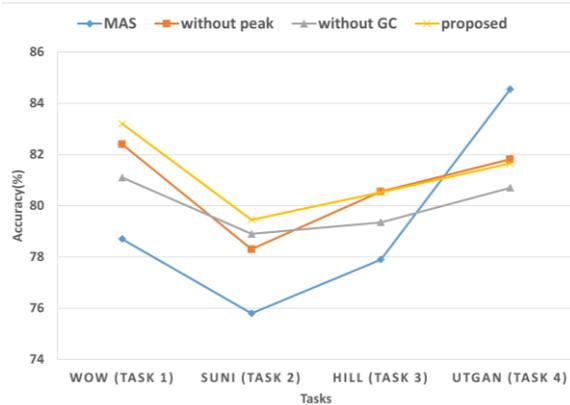

Fig. 3. Ablation analysis of the proposed curvature-based weight estimation algorithm, Peak-weight algorithm, and MAS against four sequentially steganographic methods in term of detection accuracy.

Curvature-based estimation algorithm and Peak-Weight algorithm can provide higher accuracy for steganalysis.

## 4 Conclusion

In this paper, we first introduce the continual learning scheme to steganalysis, aims to enable steganalysis model extensible for new emerging steganographic algorithms. By only retraining steganalysis model in the dataset generated by a new steganographic algorithm, making it effective against both existing steganographic algorithms and the new emerging steganographic algorithm. Specifically, we propose an APIE-based continual learning framework to mitigate the catastrophic forgetting for steganalysis. Extensive experiments demonstrate the effectiveness of our proposed scheme for steganalysis model. Moreover, due to the universality and high degree of decoupling of this framework, the proposed scheme has also great potential for other tasks with dynamic learning environments.